\documentclass{article}

\usepackage[preprint]{corl_2026} %

\usepackage{graphicx}
\usepackage{amsmath}
\usepackage{amssymb}
\usepackage{multicol}
\usepackage{booktabs}
\usepackage{subcaption}
\usepackage{lipsum}
\usepackage{cleveref}
\usepackage{float}
\usepackage{wrapfig}

\usepackage{enumitem}
\usepackage{amssymb}
\usepackage[normalem]{ulem}
\newlist{checklist}{itemize}{3}
\setlist[checklist]{label=$\square$, leftmargin=*}

\title{Sparse Autoencoders Reveal Interpretable and Steerable Features in VLA Models}

\author{
Aiden Swann$^{1}$, Lachlain McGranahan$^{3}$, Hugo Buurmeijer$^{3}$ \\ \textbf{Monroe Kennedy III$^{1,2}$, Mac Schwager$^{3}$}\\
$^{1}$Department of Mechanical Engineering, $^{2}$Department of Computer Science \\
$^{3}$Department of Aeronautics \& Astronautics\\
Stanford University \\}

\begin{document}
\maketitle
\begin{figure}[h]
    \centering
    \vspace{-5mm}
    \includegraphics[width=0.90\linewidth]{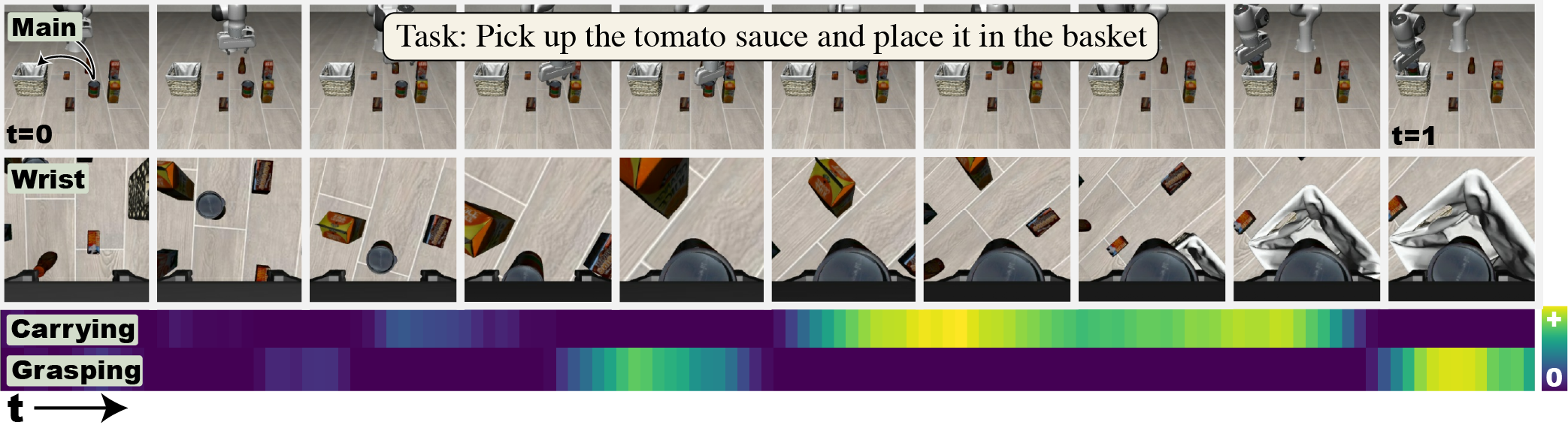}
    \caption{Activations for two general features related to grasping in the $\pi_{0.5}$ model's PaliGemma Layer 5 over an episode of LIBERO. These features activate generally for grasping and carrying behaviors across the dataset.}
    \label{fig:placeholder}
\end{figure}
\vspace{-5mm}
\begin{abstract}

Vision-Language-Action (VLA) models have emerged as a promising approach for general-purpose robot manipulation. However, little research has mechanistically explored when and why they generalize across objects, scenes, and instructions. 
To probe internal representations, we train Sparse Autoencoders (SAEs) on the VLA's hidden-layer activations. SAEs learn sparse dictionaries over model activations, often revealing features that correspond to interpretable directions in the model’s representation space. We identify SAE features corresponding to motion primitives and semantic concepts, including features that are general across episodes and causally steerable. We propose a metric to categorize features as general transferable primitives or episode-specific memorizations, offering a promising glimpse towards VLA generalization. 
We validate these findings through steering experiments on both the LIBERO simulation benchmark and on real-world DROID hardware. We find that amplifying general and semantic features induces behaviors consistent with their meanings, whereas ablating them destroys model performance. Furthermore, we demonstrate steering as a way to control behavior in unpromptable directions. Together, these results provide mechanistic evidence that VLAs can learn reusable internal features linking perception, language, and action across tasks and scenes. Our project page is located at \href{https://drvla.github.io/}{https://drvla.github.io}.

\end{abstract}
\keywords{Vision-Language-Action Models, Sparse Autoencoders, Mechanistic Interpretability, Robot Learning}

\begin{figure}[h]
    \centering
    \includegraphics[width=\textwidth]{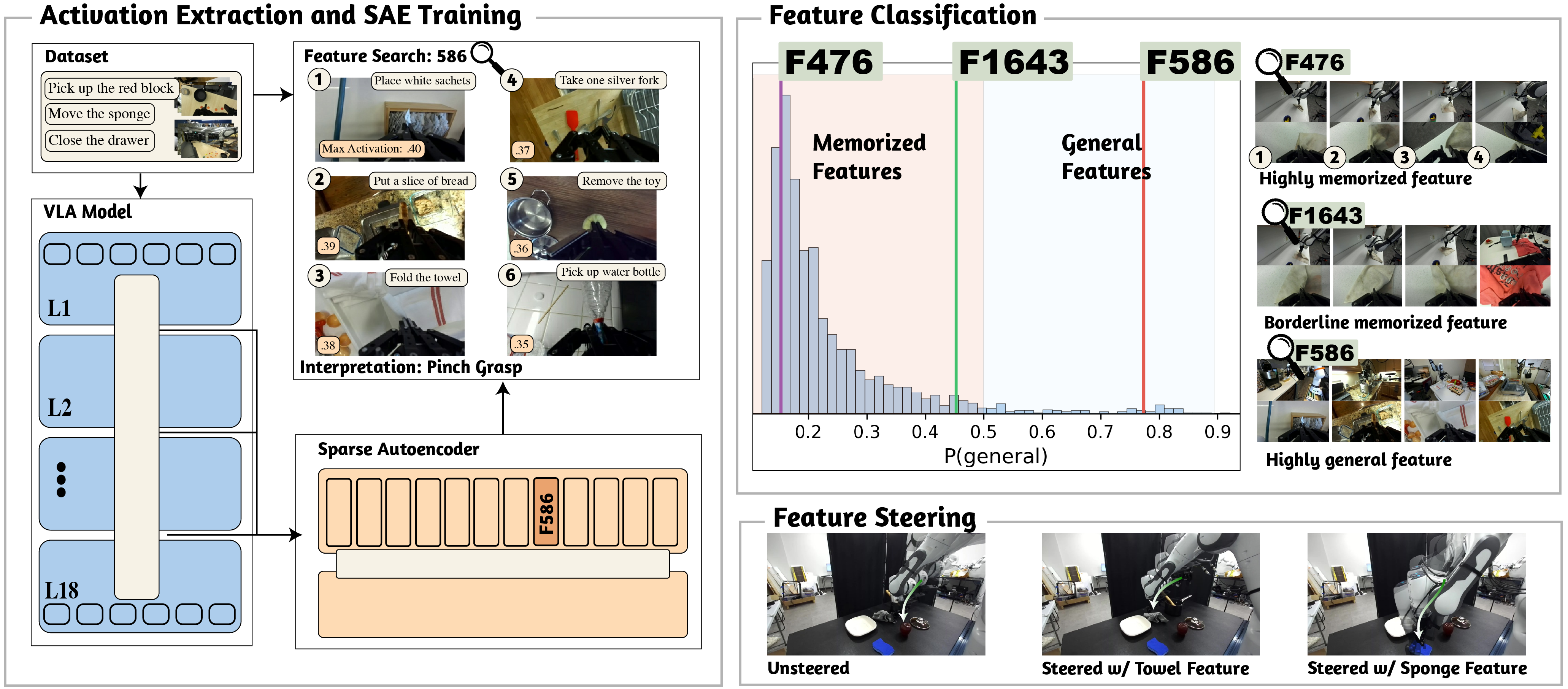}
    \caption{We collect internal activations from a VLA, train a Sparse Autoencoder (SAE), and obtain sparse features that represent both memorized and general semantic concepts. We propose metrics to categorize features by generality and offer steering results to validate our features on a DROID setup.}
    
    \label{fig:splash}
    \vspace{-20pt}
\end{figure}

\section{Introduction}

The field of robot manipulation is increasingly shaped by research in generalist policies that combine visual inputs, natural language instruction, and continuous control outputs into a single learned system. The primary example of such a policy architecture is the Vision-Language-Action (VLA) model (e.g. RT-2 \cite{RT2}, OpenVLA \cite{OpenVLA} and $\pi_0$ \cite{blackPi0}). VLA models typically couple a pretrained vision language model (VLM) backbone with a separate action decoding head. These models are pretrained on large, heterogeneous, cross-embodiment robot datasets such as OpenX Embodiment \cite{o2024open} or DROID \cite{khazatsky2024droid}.

The motivation for using VLA models is simple. Large language models (LLMs) and vision language models (VLMs) achieve impressive generalization across a wide variety of tasks \cite{floridi2020gpt, liu2023visual}. Specifically, these frontier models learn rich representations that enable generalization across text, objects, and spatial relations. VLAs attempt to leverage this widespread semantic-visual knowledge through a VLM backbone to achieve broad generalization across a variety of robot tasks in diverse visual environments, driven by open-vocabulary language prompts. However, current VLAs suffer from several pitfalls. Typically, VLAs must be fine-tuned on a specific task or embodiment to perform well. Despite rapid empirical progress on benchmarks such as LIBERO \cite{liu2023libero} and Robocasa \cite{nasiriany2024robocasa}, these models often lose language following and generalization abilities during supervised fine-tuning (SFT). Furthermore, papers like LIBERO-PRO \cite{zhou2025liberopro} have shown that models that exceed 90\% success rate under the original protocol can collapse to near-zero under systematic perturbations, implying that these policies may rely on rote memorization of action sequences and environment layouts rather than generalizing to new perceptual inputs.

We still do not have a firm grasp on how these large models work. As a result, most explanations for such brittleness are behavioral and anecdotal rather than rooted in analysis.  Specifically, we currently lack the tools to understand which parts of the model encode transferable concepts or are memorizing the data, despite being able to observe failures. Drawing inspiration from the LLM community \cite{lan2024sparseSAE, gao2024scalingSAE, kramar2026building}, this work employs mechanistic interpretability tools to understand the inner workings of learned models. Specifically, we leverage Sparse Autoencoders (SAEs), an unsupervised learning technique that disentangles superimposed representations by projecting dense activations onto a higher-dimensional sparse latent space. This technique is known for finding highly interpretable features in LLMs \cite{cunningham2023SAEhighinterp}, and we show that this trend continues with VLAs.

In this work, we apply SAEs to the residual streams of VLA models. Crucially, we verify that steering individual features during inference can predictably modulate robot behavior. Empirically, we find that SAEs uncover an interpretable basis for VLA computation, including features aligned with motion primitives, task completion, and language-relevant semantics. Our contribution is as follows: 
\begin{itemize}
    \item We propose a technique to extract \textbf{interpretable}, \textbf{general}, and \textbf{steerable} features from VLA models using Sparse Autoencoders.
    
    \item We propose several \textbf{generality quantification metrics} for SAE features using activation statistics on the fine-tuning data, without running any policy roll-outs. 
  
    \item We validate our features through extensive \textbf{simulation} and \textbf{real-world steering experiments}, showing that intervening on individual SAE features produces consistent, semantically meaningful changes in robot behavior
    
    \item We offer \textbf{Dr. VLA}, an \textbf{open-source} and user-friendly software package for \textbf{SAE training}, \textbf{feature evaluation}, and \textbf{policy steering} in VLAs (to be released).
\end{itemize}

\section{Related Work}

\subsection{Mechanistic Interpretability of VLAs}
Mechanistic Interpretability of VLAs is a nascent field with only a handful of works released in the past year. Existing work typically spans the following two goals: (i) \emph{decoding} task-relevant information from internal representations and (ii) \emph{intervening} on internal activations to causally steer behavior. On the decoding side, \citet{lu2025probing} train linear probes on OpenVLA activations to decode states such as object positions and actions. They find that probe accuracies exceed $>$90\% for most layers. This suggests that, like in LLMs, VLAs often linearly encode structured information relevant to the task, even when the end-to-end policy remains a black box. Building on this observation, \citet{molinari2025emergent} move beyond static state decoding and probe toward an emergent ``world model'' structure by testing whether state-transition vectors are recoverable from intermediate activations. While they propose SAEs as a promising next step to interpretable planning, they do not implement them.
In parallel, several other works pursue causal interventions. \citet{mech_interp_tomlin} interpret feed-forward network (FFN) directions by projecting them into token space, clustering directions by semantics (e.g., higher/lower, faster/slower), and show that manipulating the corresponding activations can steer actions. One issue is that the underlying FFN features remain highly superposed and thus may not scale cleanly to more abstract or compositional concepts without an explicit feature decomposition, such as SAEs. \citet{khan2025controlling} also apply SAEs to VLA steering, but do not assess feature interpretability or generality, rely on contrastive interventions spanning many latent directions rather than isolating individual features, train only on pretrained SAEs with no robot data, and evaluate on a single simulated gripper-opening task. Finally, \citet{buurmeijer2026observingcontrollingfeaturesvisionlanguageaction} span both observation and intervention of VLAs using linear probes. The authors find they can reliably apply control theoretic techniques to steer outputs with minimal linear interventions while operating closed loop. Furthermore, they propose leveraging SAEs to discover features without explicit labels as a direction of future work.

\section{Methods} %
\label{sec:methods}

\subsection{Sparse Autoencoders}
\label{sec:sae_auxk}
To extract interpretable features from the VLA's internal representations, we train sparse autoencoders (SAEs) on residual stream activations following the  
TopK architecture with AuxK auxiliary loss introduced by \citet{gao2024scalingSAE}. This architecture provides direct control over feature sparsity through the 
TopK activation function \citep{makhzani2015winnertakeallautoencoders} and addresses the dead latent problem through an auxiliary reconstruction loss computed over inactive    
  features \citep{gao2024scalingSAE, bricken2023towards_monosemanticity}. A detailed explanation of our methods can be found in \Cref{sec:apendix:sae_hyperparams}.

\subsection{Generality Quantification Metrics }
\label{sec:feature_metrics}

Each SAE dictionary contains many thousands of individual features. Manually inspecting all of these across many models and layers is intractable, so we define a set of per-feature activation statistics that summarize how each feature behaves within episodes and across the dataset. Together they provide a basis for distinguishing general features from memorized features, enabling automated classification at scale (\Cref{sec:results}). We summarize the metrics below at a conceptual level and use this notation throughout the paper. Full mathematical definitions and implementation choices are provided in \Cref{sec:appendix:generality_metrics}. Each feature $j$ is summarized by four activation statistics:

\textbf{Episode coverage} ($c_j$) measures how broadly a feature appears across the dataset by measuring the fraction of episodes in which the feature activates at least once. High coverage suggests that the feature is not tied to a single trajectory or scene.

\textbf{Mean onset count} ($\bar{o}_j$) measures how often a feature turns on within episodes where it is active. Features with many short activation bursts tend to correspond to repeated events or subskills, while features with few onsets are more likely to reflect sustained episode-specific context.

\textbf{Mean activation magnitude} ($\bar{a}_j$) measures the typical peak strength of a feature when it fires, averaged over active episodes. This helps distinguish weak incidental activations from strongly expressed features.

\textbf{Relative run length} ($\bar{\ell}_{r,j}$) measures how long each activation persists relative to episode length. Short, event-locked activations are more characteristic of general features, while long, sustained activations often indicate memorized scene or episode features.

Together, these metrics capture whether a feature is broad or narrow across episodes, bursty or sustained within episodes, and weakly or strongly expressed when active.

\subsection{Feature Classification}
\label{sec:feature_classification}

Using the activation statistics above, we classify SAE features as either \emph{general} or \emph{memorized}. We first manually label a small subset of features based on their top-activating trajectories, then train a logistic regression classifier on the per-feature metrics to scale this labeling across layers and models. Concretely, the classifier estimates $P(\mathrm{general}\mid m_j)$ from the metric vector $m_j = [c_j, \bar{o}_j, \bar{a}_j, \bar{\ell}_{r,j}]$. Full labeling criteria, classifier details, and validation results are provided in \Cref{sec:appendix:feature_classification}.

Following prior work on SAE interpretability, we call a feature \emph{interpretable} if its top-activating examples share a human-identifiable pattern. For VLAs, the relevant examples are observation--action trajectories rather than text sequences, so we label a feature interpretable when its temporal activation pattern consistently aligns with an identifiable sensorimotor event, visual state, or task phase.

We distinguish two categories of interpretable features. A \emph{general feature} activates across diverse episodes in response to a semantically coherent event, such as grasping, placing, task progress, or the appearance of a relevant object. These features tend to have broad episode coverage and bursty, event-locked activations. A \emph{memorized feature} activates for a narrow episode, visual scene, or task variant. These features tend to have low episode coverage and sustained activations tied to specific trajectories or environments.

This distinction is not perfectly binary: some features mix scene-specific and task-relevant structure, while others appear semantically general but are rare in the dataset. We therefore treat the classifier as a scalable tool for ranking and categorization rather than an absolute definition of feature semantics.

\subsection{Feature Steering}
\label{sec:feature_steering}
To validate the interpretation of a specific feature, we can directly steer that feature and observe the model's output. We use two complementary interventions, both targeting the decoder direction $\mathbf{v}$ of a single SAE feature. First, additive steering amplifies the feature by adding a scaled copy of $\mathbf{v}$ to the residual stream at the hooked layer,
\begin{equation}
    \mathbf{y}' = \mathbf{y} + \alpha \cdot \mathbf{v}.
\end{equation}
The scalar $\alpha$ controls the intervention strength. Second, ablative steering removes the feature by projecting the residual stream onto the orthogonal complement of $\mathbf{v}$,
\begin{equation}
    \mathbf{y}' = \mathbf{y} - (\mathbf{y}^\top \mathbf{v})\,\mathbf{v}.
    \label{eqn:project_out}
\end{equation}
Together, these allow us to probe the effect of a feature on a modeling computation. Additional implementation details are given in \Cref{sec:method:steering}.

\section{Interpretable and General Features in VLA Residual Streams} %
\label{sec:results}

We present results demonstrating that the SAE features extracted using our method are \textbf{interpretable}, \textbf{general}, and \textbf{steerable} across multiple models and datasets.

\subsection{SAEs Find Interpretable Features}
\label{sec:result:interp}

SAE features frequently have clear interpretations as visual states, task phases, and sensorimotor events. To quantify this, we randomly sample features across $\pi_{0.5}$ and OpenVLA layers and label a feature as interpretable when its temporal activation pattern, within and across episodes, consistently aligns with a human-identifiable event or state. Across 120 sampled SAE features, 95 are interpretable (79.2\%). As a baseline, we adapt the FFN-neuron analysis of \citet{mech_interp_tomlin} to our temporal activation pipeline and find that only 6/20 sampled FFN neurons are interpretable (30.0\%). While both methods can identify meaningful directions, this comparison suggests that the SAE basis yields a substantially denser set of interpretable units. Full sampling results, the FFN-neuron adaptation, and labeling details are provided in \Cref{sec:appendix:interpretability_sampling}.

\subsection{SAEs Find General Features}
\label{sec:general_features}

\begin{figure}[h]
    \centering
        \includegraphics[width=\linewidth]{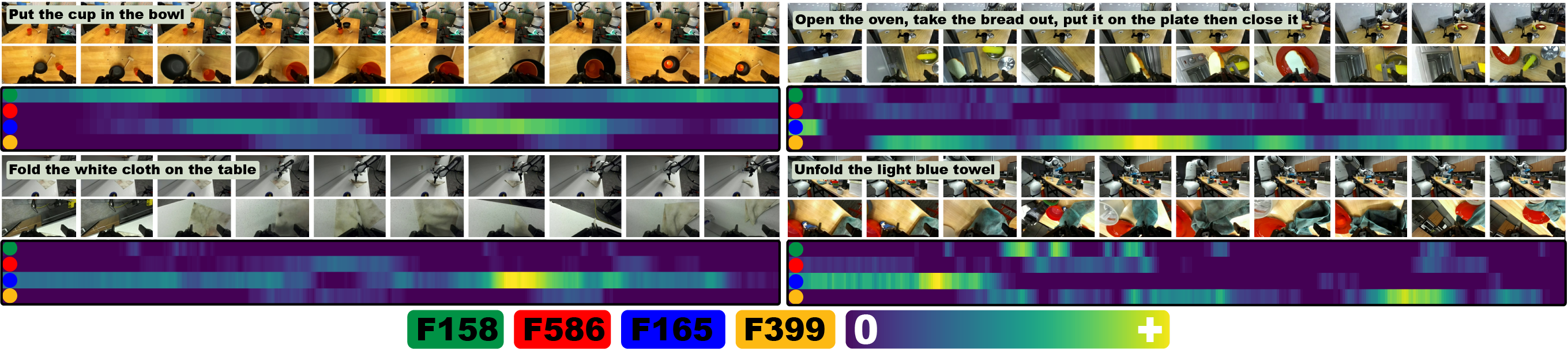}
        \caption{DROID general features across diverse tasks. For each episode, we show evenly spaced frames from the main and wrist cameras from $t_0$ to $t_f$, with per-timestep feature activations plotted below. Rows correspond to example tasks, and the color bars map to features F158, F586, F165, and F399 (legend at bottom).}
        \label{fig:droid_general_features}
        \vspace{-5pt}
\end{figure}

We find SAE features that activate consistently across diverse real-world manipulation episodes. 
\Cref{fig:droid_general_features} shows four examples from $\pi0.5$-DROID PG5 that fire across different scenes, objects, and task instructions. 
Compared with analogous LIBERO features, which we discuss in \Cref{sec:appendix:libero_general}, these DROID features activate over a substantially more diverse dataset, suggesting that they encode reusable manipulation structure rather than fixed trajectories or scenes.

The highlighted features correspond to recurring phases of manipulation. F158 activates near sub-task transitions, such as approaching, grasping, and moving toward the goal. F586 activates during precision or pinch-like grasps. F165 activates when the target object is visible between the open gripper jaws. F399 is active during grasp acquisition and placement.

\subsection{Model-Wide Automated Classification}
\label{sec:model_wide_classification}

\begin{wraptable}{l}{0.4\textwidth}
  \vspace{-10pt}
  \centering
  \small
  \begin{tabular}{lr}
  \toprule
  Model & \% General \\
  \midrule
  $\pi_{0.5}$ / LIBERO     & 2.62 \\
  $\pi_{0.5}$ / DROID      & 10.81 \\
  OpenVLA / LIBERO Goal    & 0.45 \\
  \bottomrule
  \end{tabular}
    \caption{\% of SAE features classified as general. Extended version in \Cref{tab:feature_classification_extended}}

  \label{tab:feature_classification}
  \vspace{-15pt}
\end{wraptable}

We apply the per-dataset classifiers described in \Cref{sec:feature_classification} to every SAE feature across the analyzed models and layers. \Cref{tab:feature_classification} summarizes the resulting fraction of features classified as \emph{general}.

Across all models, a small but meaningful fraction of SAE features are classified as general. The share of general features is largest in $\pi_{0.5}$-DROID, smaller in $\pi_{0.5}$-LIBERO, and smallest in OpenVLA fine-tuned on the narrower LIBERO-Goal suite. This trend suggests that broader and more diverse fine-tuning data may encourage reusable internal features, although most SAE features remain narrower or more episode-specific under our metrics. We therefore interpret the classifier as identifying a high-confidence subset of general features rather than exhaustively separating useful features from memorized ones. Classifier coefficients, validation results, and common failure modes are discussed in \Cref{sec:classifier_coefficients,sec:under_classification_generality}.

\section{Causal Validation via Feature Steering} %
\label{sec:real-results}
\subsection{Experimental Setup}
We utilize a DROID setup, which consists of a Franka Emika arm and a Robotiq 85f gripper with both a wrist and third-person camera, following \cite{khazatsky2024droid}. We utilize two scenes consisting of common objects found in both a \textit{kitchen} and \textit{office} environment, shown in \Cref{fig:real_world_scenes}. For each scene, several pick and place tasks are created with the available objects listed in \Cref{tab:real_world_tasks}. More details can be found in \Cref{sec:apendix:experimental_setup}

\subsection{Steering Validates Interpretations of General Features}

\label{sec:project_out}
Amplifying the DROID general features in \Cref{fig:droid_general_features} produces behavior consistent with their interpretations, including dwelling near sub-task transitions, earlier gripper closure, and approach--retry cycles. More details on qualitative steering results can be found in \Cref{sec:qualitative_steering_results}, and videos can be found on our \href{https://drvla.github.io}{website}.

To test whether the classifier identifies features that transfer beyond their original training episodes, we conduct ablation experiments on held-out real-world tasks. If general features encode reusable primitives, removing them should degrade performance. In contrast, features classified as memorized should be tied to specific training episodes or scenes, and therefore should have little effect on novel evaluation tasks.

We fully remove the features from the residual stream, as described in \Cref{sec:feature_steering}. This is applied for the full duration of the episode. We ablate features from three tiers of generality. For the \emph{memorized} and \emph{general} tiers, we draw four features uniformly at random from the features the classifier assigns to each category. For the \emph{most general} tier, we take the four features with the highest generality scores in the $\pi_{0.5}$ DROID PG5 dictionary. We evaluate every feature over five trials per task on two real-world tasks from \Cref{tab:real_world_tasks} (sponge, towel), with all other conditions matching, and report success rates aggregated across the four features in each tier. These particular tasks are chosen due to their high unsteered success rates.

\begin{table}[h]
\centering
\small
\begin{tabular}{lccc}
\toprule
Feature tier & Sponge & Towel & Overall \\
\midrule
Unsteered & 19/20 & 20/20 & 39/40 (97.5\%) \\
Memorized (random)   & 17/20 & 20/20 & 37/40 (92.5\%) \\
General (random)     & 10/20 & 16/20 & 26/40 (65.0\%) \\
Most general (top 4) & 0/20  & 0/20  & 0/40 (0.0\%) \\
\bottomrule
\end{tabular}
\smallskip
\caption{Feature ablation on the real-world DROID setup. For each tier, we project out four features and report success rates aggregated across them (four features $\times$ five trials $=$ 20 per task). Removing memorized features has little effect on these held-out tasks, whereas removing general features causes substantial degradation.}
\label{tab:project_out}
\end{table}

\Cref{tab:project_out} reveals a clear relationship between feature generality and the performance cost of ablation. Removing memorized features has little effect, consistent with these features being tied to training-specific episodes or scenes rather than the held-out tasks. Projecting out a random sample of general features drops success to 65.0\%, suggesting that the broader general-feature population is behaviorally important, not only a few hand-picked directions. Finally, removing the four most general features completely destroys performance, supporting the classifier's ranking. We include more detailed results on LIBERO feature ablations in \Cref{sec:libero_ablation}.

\subsection{Steering Semantic Features Biases Object Selection}
\begin{table}[h]
\centering
\begin{tabular}{l c c c c c c}
\toprule
 & \multicolumn{3}{c}{\textbf{SAE (ours)}} & \multicolumn{3}{c}{\textbf{Haon et al.}} \\
\cmidrule(lr){2-4} \cmidrule(lr){5-7}
\textbf{Task} & $\alpha$ & Success & $\Delta_{\text{base}}$ & $\alpha$ & Success & $\Delta_{\text{base}}$ \\
\midrule
Sponge & 75  & \textbf{22/25} & $+21$ & 25 & 5/25 & $+4$ \\
Towel  & 125 & \textbf{9/25}  & $+9$  & 25 & 1/25 & $+1$ \\
Toy    & 75  & \textbf{20/25} & $+17$ & 50 & 5/25 & $+2$ \\
Mug    & 125 & \textbf{13/25} & $+12$ & 25 & 5/25 & $+4$ \\
\bottomrule
\end{tabular}
\caption{Target-object grasp success, comparing SAE feature steering with FFN-neuron (Haon et al.) steering. Sponge, Towel, Toy, and Mug are steered with F1784, F8, F1727, and F39, respectively. $\Delta_{\text{base}}$ in each column reports the improvement of that method over the no-steering baseline. Full no-steering baseline distributions ($n{=}25$) --- Scene~1: 15 Apple, 8 Lid, 1 Sponge, 1 Pot; Scene~2: 20 Apple, 3 Toy, 1 Plate, 1 Mug.}
\label{tab:sae_vs_tomlin_best_alpha}
\end{table}

In these experiments, rather than providing a text instruction and then steering the policy, we give the policy an empty language instruction and test whether semantic feature steering alone can bias the model toward grasping a specific target object. This is not meant to replace or complement prompting, but simply to show that the feature directions that our SAE finds contain the object-level semantics needed to drive action selection. We compare our method against \citet{mech_interp_tomlin} and provide more details on our implementation in \Cref{sec:tomlin_implementation}. A trial is counted as a success only if the policy approaches the named target object closely enough to execute a grasp.

Without any steering, the policy collapses onto the visually dominant object in each scene. Apple captures $15/25$ trials in Scene~1 and $20/25$ in Scene~2. Against these baselines, SAE feature steering produces large, consistent shifts toward the named target ($+9$ to $+21$ over baseline), while the FFN-neuron baseline produces much smaller shifts ($+1$ to $+4$). The visualizations of the specific SAE features used for steering are available on our \href{https://drvla.github.io}{website}.

\subsection{Eliciting Behaviors That Language Prompts Cannot}
\label{sec:gripper_steering}

\begin{wraptable}{r}{0.48\textwidth}
\vspace{-10pt}
\centering
\small
\begin{tabular}{lc}
\toprule
Method & Closure \\
\midrule
\multicolumn{2}{l}{\textit{Close} ($\uparrow$)} \\
``close the gripper'' & 0.005 \\
Haon et al.           & 0.361 \\
SAE F586 (ours)       & \textbf{0.653} \\
\midrule
\multicolumn{2}{l}{\textit{Open} ($\downarrow$)} \\
``open the gripper''      & 0.575 \\
Haon et al.           & 0.938 \\
SAE F406 (ours)       & \textbf{0.510} \\
\bottomrule
\end{tabular}
\caption{Gripper closure ($0$ open, $1$ closed), averaged over the first 50 control steps and 5 trials per condition.}
\label{tab:gripper_steering}
\vspace{-10pt}
\end{wraptable}

While VLA models are generally responsive to open-world text instructions, certain behaviors can be difficult or impossible to obtain through language prompting alone. We study direct control of the gripper as a minimal test case. On the real-world DROID setup, we compare three interventions for opening and closing the gripper: (i) \emph{prompting}, where the policy is given the instruction directly; (ii) \emph{SAE steering}, where we amplify a single SAE feature; and (iii) \emph{FFN-neuron steering} following \citet{mech_interp_tomlin}. For the SAE method, we report the single best feature; for the baseline, we report the mean across the three individually steered top neurons (ranks~1--3). All steering is applied at $\alpha = 150$ on PaliGemma layer~5. We measure gripper closure on a scale where $0$ is fully open and $1$ fully closed, averaged over the first 50 control steps of each rollout and across five trials per condition.

\Cref{tab:gripper_steering} summarizes our results. Closing the gripper is unreachable through language. Instructing the policy to ``close the gripper'' leaves it almost completely open (closure $0.005$) for the roll-out. Steering recovers the behavior. Amplifying a single SAE feature, F586, the pinch-grasp feature characterized in \Cref{sec:general_features} drives the gripper to a mean closure of $0.653$. The averaged FFN baseline neuron reaches only $0.361$. Opening the gripper is the most prompt-reachable reducing closure from the towel-held starting state ($\approx 0.94$) to $0.575$. SAE steering still improves on this, with F406 reaching $0.510$. The averaged FFN baseline neuron, by contrast, barely moves the gripper from its holding state ($0.938$) and fails to release the towel.

These results show that SAE feature steering can elicit a behavior that the policy does not reliably expose through its language interface at all, and can control the gripper in both directions more effectively than either prompting or FFN-neuron steering. This reinforces our claim that SAE features provide a causal and controllable influence on VLA computation.

\section{Conclusion}
\label{sec:conclusion}
In this work, we have introduced an SAE-based mechanistic interpretability pipeline for Vision-Language-Action (VLA) models. We train sparse autoencoders on residual stream activations to obtain sparse, human-interpretable features. Across two VLA architectures and two robotics datasets, we find that many SAE features exhibit clear interpretations as motion primitives, task-progress signals, semantic representations, and episode-specific memorization. Further sampling-based analysis shows that a large fraction of learned features are interpretable across both models and layers. 

Beyond our interpretability results, we provide evidence that individual SAE directions causally influence closed-loop behavior. When we steer general features, we often observe that they modify the motion in ways consistent with their hypothesized interpretation. For example, steering a pre-grasp alignment feature causes the policy to hover above objects rather than continue grasping, while a ``sponge" feature biases the robot towards sponge grasps. Furthermore, we show that removing general features destroys policy performance. 

 These interventions are complemented by our feature-search and activation-heatmap analyses. Taken together, these provide strong evidence that SAE features capture behaviorally meaningful computations rather than purely correlational artifacts.
We introduce feature metrics (episode coverage, onset count, activation magnitude, and run length) to empirically quantify generality and support scalable feature categorization. 

While these results are not applied to model improvement in this paper, they suggest several immediate directions to make these features practically useful. One potential application of these results is during model training. Because SAEs do not require any rollouts, simulated or otherwise, they can be applied in a lightweight fashion during training. For example, the presence of episode-specific features can be used to diagnose the brittleness of common VLA fine-tuning procedures. Similarly, our feature metrics may serve as a training-time proxy for generalization without requiring real-world evaluation, which can be time-consuming and expensive.

\section{Limitations}
\label{sec:limitations}
We find that meaningful top activations of a feature does not imply reliable steerability. Specifically, we find that many \emph{clean} features exhibit limited or unpredictable causal impact when used to steer. We hypothesize that nonlinear downstream interactions (since the VLA produces actions via flow matching) or the fact that being predictive does not imply causality may inhibit steerability. Furthermore, we find that some features do not fit into the proposed general-versus-memorized feature classification. For example, some features appear highly general, activating across many different episode types, yet do so only at the end or beginning of an episode. Other features appear to be general yet activate in such a small portion of the dataset that they are classified as memorized. More details can be found in \Cref{sec:under_classification_generality}. Additionally, compared to VLM research, we use orders of magnitude less data due to data constraints specific to robotics. In our experiments, due to storage and computational constraints, we primarily used mean-pooled tokens and evaluated our SAEs on a per-timestep basis. We include some initial per-token experiments (\Cref{app:per_token_saes}), which are promising but generally less interpretable and warrant further study.

\clearpage
\acknowledgments{We would like to thank Timothy Chen for reading preliminary versions of this paper and providing feedback. We utilized the Stanford Marlowe cluster for several experiments \cite{marlowe}. Aiden Swann is supported by NSF GRFP Fellowship No. DGE-2146755.}

\bibliography{references}

\clearpage
\appendix
\crefalias{section}{appendix}
\crefalias{subsection}{appendix}
\crefalias{subsubsection}{appendix}

\section{Models, Datasets, and Activation Collection}
\label{app:models_datasets_activations}

\subsection{Models}
\label{app:models}

We perform our analysis on three VLA models spanning two architectures and robot embodiments. Our primary model is $\pi_{0.5}$ \cite{intelligence2025pi05visionlanguageactionmodelopenworld}, a state-of-the-art VLA that couples a PaliGemma \cite{beyer2024paligemma} vision-language backbone (Gemma 2B \cite{team2024gemma}, 18 transformer layers, $d = 2048$, 400M SigLIP image encoder) with a dedicated action expert (300M, 18 transformer layers, $d = 1024$). The PaliGemma (PG) backbone processes three camera images (base, left wrist, right wrist); however, for our cases, we use only two input images, along with a natural-language instruction. The action expert (AE) decodes these representations into continuous robot actions via iterative denoising, attending to the PaliGemma key-value cache. We use two fine-tuned variants open-sourced by Pi: $\pi_{0.5}$-LIBERO and $\pi_{0.5}$-DROID, both fine-tuned on the respective datasets with a technique known as knowledge insulation \cite{driess2025knowledgeinsulatingvisionlanguageactionmodels}. We choose both because LIBERO provides a controlled simulation environment for closed-loop evaluations, while DROID offers some real-world diversity, enabling us to identify general semantic features.

As a secondary model, we study OpenVLA \cite{OpenVLA}, an open-source VLA built on a Llama~2 7B backbone (32 transformer layers, $d = 4096$). We use the publicly available checkpoint fine-tuned on the LIBERO Spatial suite (\texttt{openvla/openvla-7b-finetuned-libero-spatial}). OpenVLA uses a single language model for both perception and action prediction, predicting actions autoregressively. We include OpenVLA primarily to test whether the features and phenomena we observe in $\pi_{0.5}$ generalize across VLA architectures with fundamentally different action decoding mechanisms.

\begin{table}[h]
\centering
\small
\begin{tabular}{lrrrr}
\toprule
Dataset & Episodes & Tasks & Timesteps \\
\midrule
LIBERO & 1,693 & 40 & 273,465 \\
DROID (2k subset) & 2,000 & 1,545 & 567,088 \\
LIBERO-Goal (OpenVLA) & 428 & 10 & 52,042 \\
\bottomrule
\end{tabular}
\smallskip
\caption{Dataset statistics for the models we analyze in this work.}
\label{tab:datasets}
\end{table}
\subsection{Datasets}
\label{sec:method:dataset}

LIBERO \cite{liu2023libero} is a simulated benchmark covering a limited set of tabletop tasks built on the Robosuite simulator. We collect activations from the full training set provided via the HuggingFace repository (\texttt{physical-intelligence/libero}) used by the $\pi_{0.5}$ training pipeline. This comprises 1,693 episodes spanning 40 unique tasks across four suites: LIBERO-Spatial (10 tasks, 432 episodes), LIBERO-Object (10 tasks, 454 episodes), LIBERO-Goal (10 tasks, 428 episodes), and LIBERO-10 (10 tasks, 379 episodes). The dataset contains 273,465 total timesteps with 7-dimensional actions (6-DoF end-effector + gripper). Notably, the Spatial and Object suites each share a single base scene across all 10 tasks therefore the dataset contains fewer visual environments (${\sim}$20) than unique task descriptions, and considerable similarities between episodes as noted in \cite{zhou2025liberopro}.

DROID~\cite{khazatsky2024droid} is a large-scale, in-the-wild robot manipulation dataset collected across diverse real-world environments on Franka Panda robots. We collect activations on a randomly sampled subset of 2,000 episodes (1,750 successful, 250 failed) drawn from the RLDS-formatted DROID corpus. We intentionally consider failed episodes, as they represent a traditional underexamined form of robotic data. This subset spans 1,545 unique task instructions and 567,088 total timesteps with 8-dimensional actions (7 joint positions + gripper). Note that $\pi_{0.5}$-DROID was trained by Pi on the full DROID dataset (${\sim}$75k episodes); our 2,000-episode subset is used solely for activation collection and analysis.

For OpenVLA analysis, we collect activations only from the LIBERO-Goal suite, matching the model's fine-tuning distribution. This subset contains
428 episodes across 10 goal-conditioned tasks with 52,042 total
timesteps. 

\subsection{Activation Collection}
\label{app:activation_collection}
For each model, we collect residual-stream activations from the output of the full transformer block, after both the self-attention and MLP sublayers and their residual connections. This is the standard residual-stream location used in prior work on SAE-based mechanistic interpretability \cite{gao2024scalingSAE}. Activations are captured with PyTorch \texttt{register\_forward\_hook} callbacks on the target decoder-layer modules.

At each timestep, the $\pi_{0.5}$ PaliGemma backbone processes a variable-length sequence of tokens: three 256-token image embeddings (one per camera, yielding 768 image tokens) plus additional instruction and state tokens. By default, we mean-pool all tokens within the hook into a single $d$-dimensional vector per timestep. Thus, each episode produces an activation matrix of shape $(T,d)$, where $T$ is the number of timesteps. All main-text SAE results use these mean-pooled timestep activations unless otherwise noted.

We use mean-pooled activations because the timestep is the natural unit of analysis for robot behavior and because storing all token activations would be prohibitively expensive. For example, collecting all image-patch token activations for a single layer over the 2,000-episode DROID subset would require several terabytes of storage. We also collect a smaller set of per-token activations for exploratory analysis by pooling each camera's image patches into one image token while preserving individual text-token activations. At each timestep, this yields 2 image vectors (one per camera, each obtained by mean-pooling that camera’s 256 patch
tokens) and $\sim$21 individual text-token vectors, all of dimension 2048. These per-token SAE results are presented in \Cref{app:per_token_saes}.

\section{Sparse Autoencoder Training}
\label{app:sae_training}

\subsection{SAE Architecture and Training Hyperparameters}
\label{sec:apendix:sae_hyperparams}
We train all SAEs using the TopK architecture with AuxK auxiliary loss \citet{gao2024scalingSAE}, with default hyperparameters summarized in \Cref{tab:sae_hyperparams}. Given an input activation $\mathbf{x} \in \mathbb{R}^d$, the SAE first applies per-sample normalization. A learned pre-bias $\mathbf{b}_{\mathrm{pre}}$, initialized to the geometric median of training activations, is subtracted, followed by per-sample mean subtraction and $\ell_2$ normalization:
\begin{equation}
    \tilde{\mathbf{x}} =
    \frac{(\mathbf{x} - \mathbf{b}_{\mathrm{pre}}) - \mu}
    {\lVert(\mathbf{x} - \mathbf{b}_{\mathrm{pre}}) - \mu\rVert_2},
\end{equation}
where $\mu$ is the scalar mean over the model dimension. The normalized input is encoded into a sparse representation:
\begin{equation}
    \mathbf{z} =
    \mathrm{ReLU}\!\left(\mathrm{TopK}\!\left(\mathbf{W}_{\mathrm{enc}}\tilde{\mathbf{x}}\right)\right),
\end{equation}
where TopK retains the $k$ largest pre-activation values and zeros the rest. The reconstruction in normalized space is $\hat{\tilde{\mathbf{x}}}=\mathbf{W}_{\mathrm{dec}}\mathbf{z}$, which is then un-normalized by restoring the saved $\ell_2$ norm, mean, and pre-bias. Decoder columns are constrained to unit norm, so each feature's contribution is controlled by its scalar activation coefficient. Neither the encoder nor decoder uses bias terms.

The total loss combines normalized reconstruction error with the AuxK auxiliary loss:
\begin{equation}
    \mathcal{L}
    =
    \frac{\lVert \mathbf{x} - \hat{\mathbf{x}} \rVert_2^2}{C_{\mathrm{MSE}}}
    +
    \alpha
    \frac{\lVert \tilde{\mathbf{e}} - \hat{\tilde{\mathbf{e}}}_{\mathrm{aux}} \rVert_2^2}
    {C_{\mathrm{MSE}}},
\end{equation}
where $C_{\mathrm{MSE}}$ is the variance of centered activations computed at initialization, $\alpha=1/32$, $\tilde{\mathbf{e}}=\tilde{\mathbf{x}}-\hat{\tilde{\mathbf{x}}}$ is the normalized reconstruction residual, and $\hat{\tilde{\mathbf{e}}}_{\mathrm{aux}}$ is the auxiliary reconstruction from the top-$k_{\mathrm{aux}}$ dead latents applied to this residual. A latent is considered dead if it has not activated within the last 500 optimization steps. Encoder weights are initialized as a scaled transpose of the decoder, $\mathbf{W}_{\mathrm{enc}}=\mathbf{W}_{\mathrm{dec}}^\top\sqrt{k/n}$, where $n$ is the residual-stream dimension. Decoder gradients are projected onto the tangent plane of the unit-norm constraint \citep{bricken2023towards_monosemanticity}, and gradient norms are clipped at 1.0.

\begin{table}[h]
\centering
\small
\begin{tabular}{lr}
\toprule
\textbf{Hyperparameter} & \textbf{Value} \\
\midrule
Expansion ratio & 1 \\
Active features $k$ & 100 \\
Auxiliary $k_{\mathrm{aux}}$ & 512 \\
Auxiliary loss coefficient $\lambda_{\mathrm{aux}}$ & $1/32$ \\
Learning rate $\eta$ & $1 \times 10^{-4}$ \\
Optimizer & Adam ($\beta_1=0.9$, $\beta_2=0.999$) \\
Batch size & 4096 \\
Training epochs & 100 \\
Geometric median samples & 10{,}000 \\
\bottomrule
\end{tabular}
\smallskip
\caption{Default SAE training hyperparameters. An expansion ratio of $ER=1$ means the SAE hidden dimension equals the input dimension $d$, e.g. 2048 features for PaliGemma layers. Expansion-ratio ablations are shown in \Cref{fig:training_ablations}.}
\label{tab:sae_hyperparams}
\end{table}

We train SAEs with a $1\times$ expansion ratio on the $\pi_{0.5}$ layers described in \Cref{app:activation_collection}: PaliGemma layers 0, 5, 11, and 17 ($d=2048$, $k=100$), and action expert layers 0, 5, 11, and 17 ($d=1024$, $k=64$). We also train SAEs on the selected OpenVLA layers. For OpenVLA, we use $ER=0.5$ to keep the dictionary size comparable to the 2048-feature $\pi_{0.5}$ PaliGemma SAEs. All other hyperparameters are unchanged.

\begin{figure}[t]
    \centering
    \includegraphics[width=0.65\linewidth, trim=0 0 0 30pt, clip]{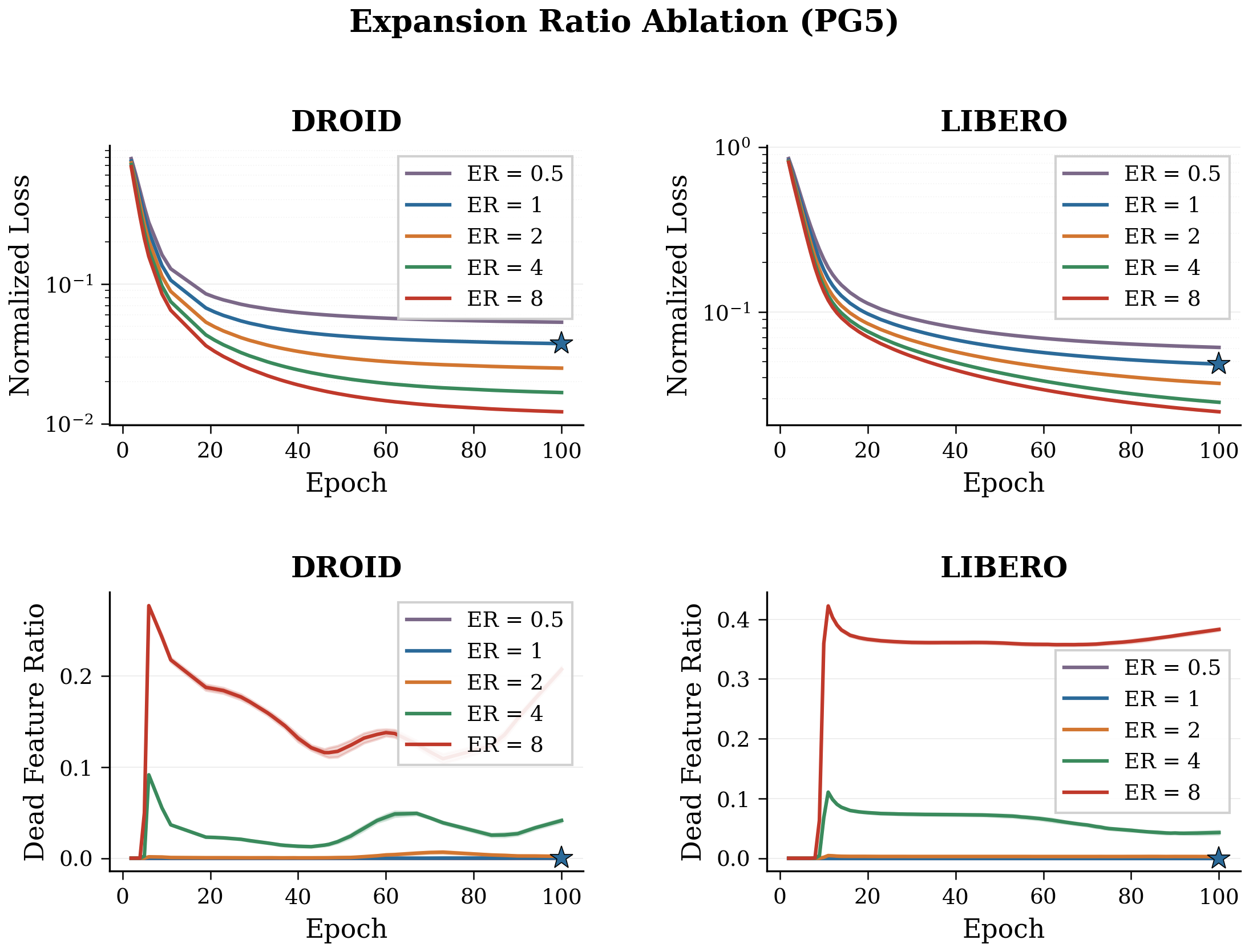}
    \caption{Ablations over SAE expansion ratio and learning rate on $\pi_{0.5}$ PG5 activations. Top row: reconstruction loss versus epoch for DROID (left) and LIBERO (right). Bottom row: dead feature ratio versus epoch. The star marks the setting used throughout this paper, $ER=1$, chosen as a trade-off between reconstruction loss and dead feature ratio.}
    \label{fig:training_ablations}
\end{figure}

A notable departure from typical VLM interpretability work is our low expansion ratio. As shown in \Cref{fig:training_ablations}, larger expansion ratios lead to substantially more dead features while providing similar interpretability in our setting. We hypothesize this is likely due to the much smaller scale datasets sizes which we typically work with in robotics. We therefore use $ER=1$ as a compact dictionary size that balances reconstruction quality and feature utilization.

\subsection{Multi-Seed Ablation}
\label{sec:apendix:random_seed}
\begin{figure}[h]
    \centering
    \includegraphics[width=0.5\linewidth]{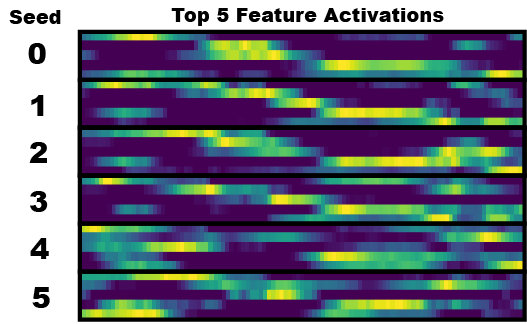}
    \caption{Activations from SAEs trained on $\pi_{0.5}$ LIBERO from 6 randomly initialized seeds. We plot the top five features for episode 689; activations are normalized for visualization and ordered by consistency relative to seed 0. The shared patterns across random initializations provide evidence that the learned SAE features arise from the underlying model.}
    \label{fig:random_seed}
\end{figure}

To verify that the features we identify reflect structure in the model's representations rather than imaginations of the SAE, we train six SAEs from independent random seeds on the same activation data ($\pi_{0.5}$ LIBERO PG5). \Cref{fig:random_seed} shows that the top features for a given episode are consistently recovered across seeds, with similar temporal activation patterns. 

\section{Feature Metrics and Generality Classification}
\label{app:feature_metrics_classification}

\subsection{Generality Metrics}
\label{sec:appendix:generality_metrics}

Let $f_j(\mathbf{x}_t^{(e)}) \in \mathbb{R}_{\geq 0}$ denote the activation of SAE feature $j$ at timestep $t$ in episode $e$, and let $T^{(e)}$ be the number of timesteps in episode $e$. Let $E = \{e_1, e_2, \ldots, e_N\}$ denote the set of all episodes in the fine-tuning dataset and  $E_j^+ = \{e : \exists\, t,\; f_j(\mathbf{x}_t^{(e)}) > 0\}$ be the set of episodes in which feature $j$ fires at least once.

\paragraph{Episode Coverage.} Episode coverage is the fraction of episodes in the fine-tuning dataset where the feature activates at least once:
\begin{equation}
    c_j = \frac{|E_j^+|}{|E|}
    \label{eqn:episode_coverage}.
\end{equation}  
Higher episode coverage indicates the feature is active across diverse tasks and therefore has greater generality.

\paragraph{Mean Onset Count.} An onset of a feature is the transition of that feature from inactive to active. Rather than defining an active feature as simply $f_j(\mathbf{x}_t^{(e)}) \neq 0$ for a specific timestep, we instead add an activation threshold $\tau_{\mathrm{on}}$ to suppress the effects of noise. Each feature $j$ maintains a binary state $s$ with $s_0 = 0$. The state transitions are:
\begin{equation}
s_t = \begin{cases} 
1 & \text{if } f_j(x_t) > \tau_{\mathrm{on}} \\ 
0 & \text{if } f_j (x_t) = 0 \\ 
s_{t-1} & \text{otherwise} 
\end{cases},
\end{equation}
with activation threshold $\tau_{\mathrm{on}} = 0.1$.  An onset is counted at each $0 \to 1$ transition, giving the per-episode onset count:
\begin{equation}
o_j = \sum_{t=1}^{T} \max(0, s_t - s_{t-1}).
\end{equation}

The mean onset count averages over active episodes only, decoupling it from episode coverage:
\begin{equation}
\bar{o}_j = \frac{1}{|E_j^+|} \sum_{e \in E_j^+} o_j^{(e)}.
\end{equation}
Since every active episode has at least one onset, $\bar{o}_j \geq 1$ for any feature with $c_j > 0$.
General features typically exhibit $\bar{o}_j \gg 1$, indicating bursty, event-driven activation.

\paragraph{Mean Activation Magnitude.} For each active episode $e \in E_j^+$, we record the maximum activation of feature $j$ across all timesteps. The mean activation magnitude averages these per-episode maxima:
\begin{equation}
\bar{a}_j = \frac{1}{|E_j^+|} \sum_{e \in E_j^+} \max_{t} f_j(\mathbf{x}_t^{(e)}).
\end{equation}

This metric captures the typical peak intensity of a feature when it fires, averaged across all episodes in which it is active.

\paragraph{Relative Run Length.} For each active episode, the run length is the mean number of consecutive active timesteps per onset:
\begin{equation}
r_j = \frac{1}{o_j} \sum_{t=1}^{T} s_t.
\end{equation}
Normalizing by episode length yields the relative run length, which expresses the average activation duration as a fraction of the episode:
\begin{equation}
\bar{\ell}_{r,j} = \frac{1}{|E_j^+|} \sum_{e \in E_j^+} \frac{r_j^{(e)}}{T^{(e)}}.
\end{equation}

Values near $0$ indicate brief, transient activations; values near $1$ indicate sustained activation across the full episode.
General features tend to have low $\bar{\ell}_r$ (bursty), while memorized features often exhibit high $\bar{\ell}_r$ (sustained).

\subsection{Manually Labeling Features}
\label{sec:appendix:feature_classification}
We manually label a subset of features and use these to train an automatic classifier. We label these features through a three-stage visual inspection pipeline that examines activation patterns at the episode, cross-episode, and dataset levels. 

\textbf{Stage 1: Episode-Level Screening.} Starting from a randomly selected episode, we examine a feature activation heatmap that displays the top 50--100 features at a given layer. Each row of the heatmap shows the activation magnitude of a single feature across all timesteps, aligned with the corresponding camera frames, as shown in \Cref{fig:general_features_libero}. We refer to this visualization as the \emph{Activation Viewer}. We identify candidate general features by their \emph{bursty} activation signature, meaning short, high-magnitude peaks coinciding with identifiable behavioral events (e.g. grasps, placements, object appearances). We identify candidate memorized features by their \emph{sustained} signature, meaning near-uniform activation across most time steps without alignment with behavioral transitions. 

\textbf{Stage 2: Cross-Episode Validation.} For each candidate, we query a precomputed index over all SAE activations in the dataset, which we call the \emph{Feature Search} index. This returns the globally top-activating timesteps and the top-10 diverse episodes ranked by maximum activation. General features should exhibit peak activations spanning multiple episodes and tasks with consistent, even alignment; memorized features should show activations concentrated in one or a few episodes, with a steep drop-off in magnitude. We additionally verify that this new episode subset exhibits the same per-episode temporal pattern in the Activation Viewer (bursty vs.\ sustained), ensuring consistency with the original randomly selected episode.

\textbf{Stage 3: Labeling Criteria.} A feature receives the label \emph{general} if: (1) activation bursts align with semantically consistent events across the top diverse episodes, (2) the activation pattern reproduces in held-out episodes viewed in the \textit{Activation Viewer}, and (3) the global per-feature metrics (mean onset count, mean active activation, episode coverage) indicate this feature is persistent across a large portion of the training dataset. A feature receives the label \emph{memorized} if: (1) activation is sustained within the top episode(s), (2) the top episodes cluster around $\leq 2$ visual scenes, and (3) the global per-feature metrics indicate this feature activates on a small subset of the training dataset. If there is any ambiguity in the interpretation of the \textit{Feature Search} results, then the given feature is excluded.
\subsection{Automated Classification}
\label{sec:automated_classification}                    

Manual labeling is precise but does not scale to thousands of features across multiple models and layers. To classify features at scale, we fit a logistic regression classifier to the four temporal metrics defined in \Cref{sec:feature_metrics}, using 30 manually labeled features (15 general, 15 memorized) from a single reference layer as training data. The classifier takes the form:                                       
  \begin{equation}                        
      P(\text{general} \mid \mathbf{m}) = \sigma\!\left(\beta_0 + \beta_1 \bar{o} + \beta_2 c + \beta_3 \bar{a} +   
  \beta_4 \bar{\ell}_r\right),
  \end{equation}
where $\bar{o}$ is mean onset count, $c$ is episode coverage, $\bar{a}$ is mean activation magnitude, $\bar{\ell}_r$ is relative run length, and $\sigma$ is the logistic sigmoid. The classifier operates on unnormalized metric values, so the learned decision boundary can be applied across layers within the same model without per-layer normalization. We train one classifier per fine-tuning dataset (LIBERO and DROID) and apply it to all models trained on that dataset. The OpenVLA classifier reuses the LIBERO boundary, since OpenVLA is also fine-tuned on LIBERO.

\subsection{Classifier Coefficients}
\label{sec:classifier_coefficients}

\begin{table}[b]                        
  \centering                              
  \small                                  
  \begin{tabular}{llrrrr}                 
  \toprule                                
  Model & Layer(s) & \# Features & \# General & \# Memorized & \% General \\
  \midrule
  \multicolumn{6}{l}{\textit{$\pi_{0.5}$ -- LIBERO}} \\
   & PG5 & 2044 & 32 & 2012 & 1.57\%\\
   & PG avg (0, 5, 11, 17) & 7175 & 188 & 6987 & 2.62\% \\
  \midrule
  \multicolumn{6}{l}{\textit{$\pi_{0.5}$ -- DROID}} \\
   & PG5 & 2046 & 104 & 1942 & 5.08\%\\
   & PG avg (0, 5, 11, 17) & 6649 & 719 & 5930 & 10.81\% \\
  \midrule
  \multicolumn{6}{l}{\textit{OpenVLA -- LIBERO Goal}} \\
   & Layer 8 & 1775 & 8 & 1767 & 0.45\%\\
   & LM avg (0, 8, 16, 24, 31) & 9389 & 42 & 9347 & 0.45\% \\
  \bottomrule
  \end{tabular}
  \smallskip
  \caption{Feature classification results across models and layers. We report the number and percentage of SAE features classified as general by the activation-pattern classifier. The remaining features are labeled memorized, reflecting features that are narrower, more episode-specific, or not captured by our generality criteria. Individual layers serve as representative baselines; averages are computed across all analyzed layers within each subnetwork.}
  \label{tab:feature_classification_extended}
\end{table}

We fit one logistic regression classifier per fine-tuning dataset using the activation statistics defined in \Cref{sec:appendix:generality_metrics}.

For LIBERO, the fitted coefficients are
\[
\beta_0 = -4.20,\quad
\beta_c = 1.80,\quad
\beta_{\bar{o}} = 1.89,\quad
\beta_{\bar{a}} = 0.52,\quad
\beta_{\bar{\ell}_r} = -0.36,
\]
achieving 100\% leave-one-out cross-validation accuracy on the 30 manually labeled examples. Mean onset count and episode coverage contribute nearly equally, suggesting that in LIBERO both bursty event-aligned activation and broad episode presence are needed to distinguish general features from narrower, scene-specific features. The negative coefficient on relative run length indicates that sustained activations are less associated with general features under this classifier. Because OpenVLA is also fine-tuned on LIBERO, we apply the LIBERO classifier to its feature dictionaries.

For DROID, the fitted coefficients are
\[
\beta_0 = -1.78,\quad
\beta_c = 2.36,\quad
\beta_{\bar{o}} = 0.74,\quad
\beta_{\bar{a}} = 0.35,\quad
\beta_{\bar{\ell}_r} = -1.04,
\]
achieving 96.7\% leave-one-out cross-validation accuracy. Compared with LIBERO, episode coverage receives a larger positive weight, consistent with DROID's greater scene and task diversity: a feature that activates across many DROID episodes is less likely to be tied to a single repeated scene or trajectory. DROID also assigns a stronger negative weight to relative run length, suggesting that long, sustained activations are especially indicative of non-general features in longer and more variable real-world episodes.

Overall, the learned coefficients align with the qualitative distinction used in our manual labels: general features tend to be broad across episodes, bursty within episodes, and event-aligned, while non-general features tend to be narrower and more sustained. We emphasize, however, that this classifier is intended as a scalable heuristic for ranking and categorizing features, not as an absolute semantic boundary.

\subsection{Under-classification of Generality}
\label{sec:under_classification_generality}

These classification results likely underestimate the number of general features within a given layer, especially in high-diversity datasets such as DROID. Two examples illustrate a class of features that our metrics fail to capture: F1939 from LIBERO PG5 and F1381 from DROID PG5. 

\paragraph{F1939 (LIBERO PG5):}$\bar{o} = 1.00$, $c = 0.732$, $\bar{a} = 0.037$, $\bar{\ell}_r = 0.127$.  Despite an episode coverage of 73\%, the onset count of exactly 1.0 causes the classifier to label this feature as memorized.  All activations occur within the first 20 timesteps of each episode regardless of scene or task, consistent with encoding the robot ``home'' position---a pose shared across all LIBERO environments.  The feature is general by visual inspection (scene- and task-invariant, present in nearly three-quarters of episodes), but its single-onset-per-episode pattern yields an $\bar{o}$ indistinguishable from a memorized feature that fires once in a narrow set of episodes.

\paragraph{F1381 (DROID PG5):} $\bar{o} = 1.00$, $c = 0.226$, $\bar{a} = 0.065$, $\bar{\ell}_r = 0.990$.  This feature is activated during the grasp of a lid, regardless of lid type (metal pot lid, glass lid, paper cup lid, green plastic lid) and the scene. It fires in 116 of the 135 episodes whose task instructions contain the word ``lid'' (86\% recall within that subset), and in 336 additional episodes with instructions such as ``put the pot on the right side of the table and open it,'' which are accomplished via lid grasps.  However, because lid-related episodes constitute only 6.7\% of the DROID dataset, the episode coverage of 0.226 falls below the classifier's decision boundary.  The near-unity relative run length ($\bar{\ell}_r = 0.990$) further penalizes it, as the feature remains active throughout the lid-grasp episodes rather than activating in brief bursts.

Both cases expose the same structural limitation: features that activate once per episode ($\bar{o} \approx 1$) across a semantically coherent but proportionally small subset of the dataset will be classified as memorized.  The classifier conflates low burstiness with memorization because, in the labeled training set, general features are predominantly multi-onset.  Addressing this would require either a dataset-diversity-aware normalization of episode coverage or an additional metric that captures cross-scene consistency independently of activation frequency.
\section{Interpretability Sampling and FFN Neuron Baseline}
\label{sec:appendix:interpretability_sampling}

We estimate the fraction of interpretable features using a sampling-based protocol. For each SAE dictionary listed in \Cref{tab:interpretability_sampling}, we randomly sample 20 features and label each feature as interpretable or non-interpretable. A feature is labeled \emph{interpretable} if its temporal activation pattern, both within and across episodes, consistently aligns with an identifiable sensorimotor event, visual state, object configuration, or task phase. This criterion is distinct from our general-versus-memorized classification: a feature may be interpretable while still being narrow or episode-specific.

\begin{table}[h]
\centering
\small
\begin{tabular}{lllrrrr}
\toprule
Method & Model & Dataset / Layer & $n$ sampled & \# interp. & \# non-interp. & \% interp. \\
\midrule
SAE & $\pi_{0.5}$ & LIBERO PG5  & 20 & 18 & 2 & 90\% \\
SAE & $\pi_{0.5}$ & LIBERO PG11 & 20 & 16 & 4 & 80\% \\
SAE & $\pi_{0.5}$ & DROID PG5  & 20 & 17 & 3 & 85\% \\
SAE & $\pi_{0.5}$ & DROID PG11 & 20 & 14 & 6 & 70\% \\
SAE & OpenVLA & LIBERO Goal Layer 8 & 20 & 16 & $4^{*}$ & 80\% \\
SAE & OpenVLA & LIBERO Goal Layer 24 & 20 & 14 & 6 & 70\% \\
\midrule
\textbf{SAE Total} &  &  & \textbf{120} & \textbf{95} & \textbf{25} & \textbf{79.2\%} \\
\midrule
FFN neuron & $\pi_{0.5}$ & DROID PG5 & 20 & 6 & 14 & 30\% \\
\bottomrule
\multicolumn{7}{l}{\footnotesize $\ast$ Two of the four non-interpretable features were inactive.}
\end{tabular}
\smallskip
\caption{SAE features are frequently interpretable across models, datasets, and layers. As a baseline, we adapt the FFN-neuron approach of \citet{mech_interp_tomlin} to our temporal activation-based feature search pipeline and apply the same 20-feature sampling and labeling protocol. The feature IDs of the random samples for all models are available for cross-referencing on our \href{https://drvla.github.io}{website}.}
\label{tab:interpretability_sampling}
\end{table}

\subsection{Adapting the FFN-neuron baseline.}
The method of \citet{mech_interp_tomlin} treats individual FFN neurons as candidate steering directions. We adapt their method at the level of the unit of analysis: each FFN neuron is treated as one candidate feature and passed through the same activation-space analysis used for SAE features.

We then pass the resulting activations through the same Feature Search pipeline used for SAE features. For each neuron, we compute the same activation statistics, retrieve top-activating timesteps and top diverse episodes, and visualize temporal activations in the same Activation Viewer. Manual labeling follows the same interpretability criterion used for SAE features. Of the 16,384 FFN neurons, 16,103 fire on at least one timestep in the DROID subset, while 281 are inactive.

Using this matched pipeline, only 6/20 sampled FFN neurons are labeled as interpretable, compared with 95/120 sampled SAE features. This comparison isolates the difference between the native FFN neuron basis and the SAE-learned feature basis while holding the activation collection, indexing, visualization, and labeling procedures fixed. The result suggests that SAEs provide a substantially more interpretable basis for temporal VLA behavior than raw FFN neurons, although FFN neurons can still contain meaningful steerable directions.

\section{Additional Feature Visualizations}
\label{app:additional_feature_visualizations}

\subsection{LIBERO General Features}
\label{sec:appendix:libero_general}

\begin{figure}[h]
    \centering
    \includegraphics[width=\linewidth]{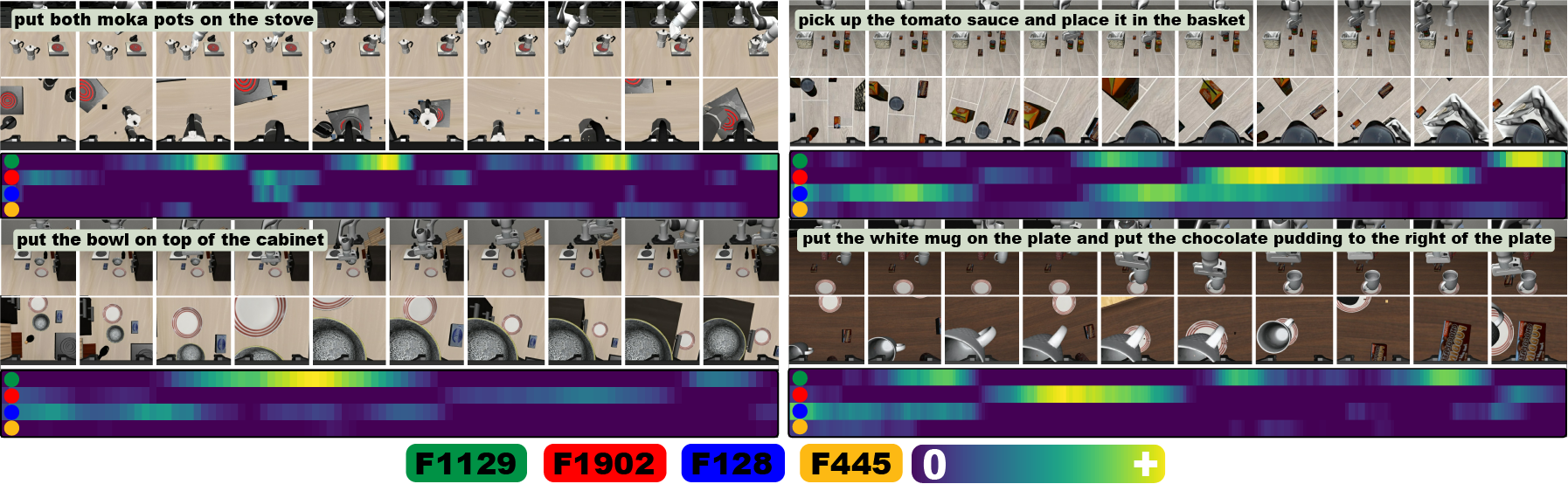}
    \caption{LIBERO general features across multiple tasks. Each row shows evenly spaced wrist- and main-camera frames from one episode, with per-timestep activations for four SAE features below. The highlighted features activate consistently across scenes, goal objects, and numbers of grasps.}
    \label{fig:general_features_libero}
\end{figure}

We observe analogous general features in $\pi0.5$-LIBERO PG5. The four features in \Cref{fig:general_features_libero} all have episode coverage $>0.99$ and activate across changes in scene, goal object, and number of grasp subgoals. F1129 activates around object grasp and placement. F1902 activates during object transport. F128 activates during pre-grasp alignment when the end-effector is positioned above the target object. F445 activates near goal placement and task completion. These features provide a counterpart to the DROID features discussed in the main text, demonstrating that SAEs recover a reusable manipulation structure across both simulated and real-world VLA datasets. Interactive top-activating episodes are available on our \href{https://drvla.github.io}{website}.

\subsection{Cross-Layer Feature Visualization}

\begin{figure}[h]
    \centering
    \begin{subfigure}[t]{0.32\linewidth}
        \centering
        \includegraphics[width=\linewidth]{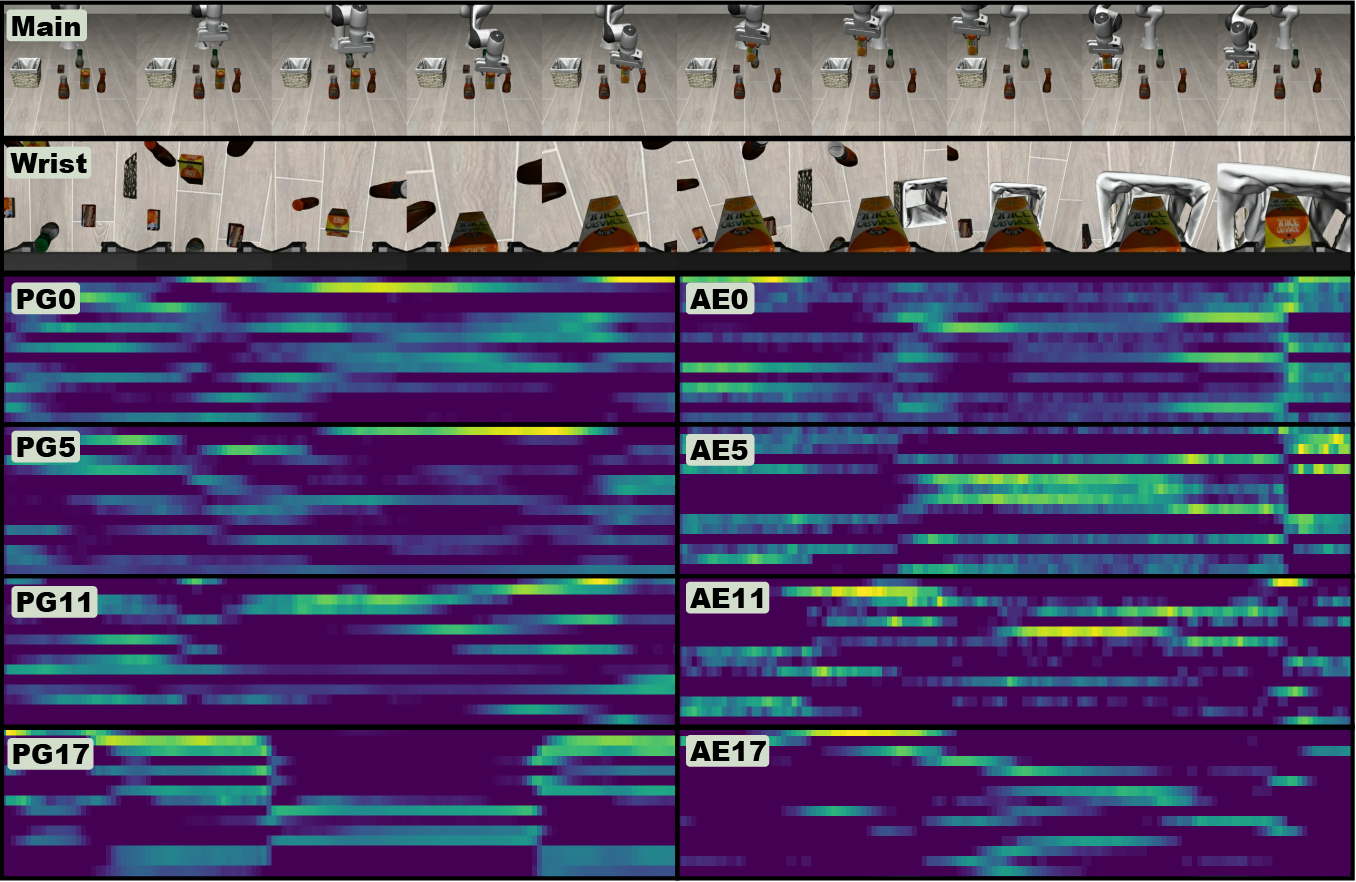}
        \caption{LIBERO episode 870: \textit{pick up the orange juice and place it in the basket}.}
        \label{fig:top15_libero}
    \end{subfigure}
    \hfill
    \begin{subfigure}[t]{0.32\linewidth}
        \centering
        \includegraphics[width=\linewidth]{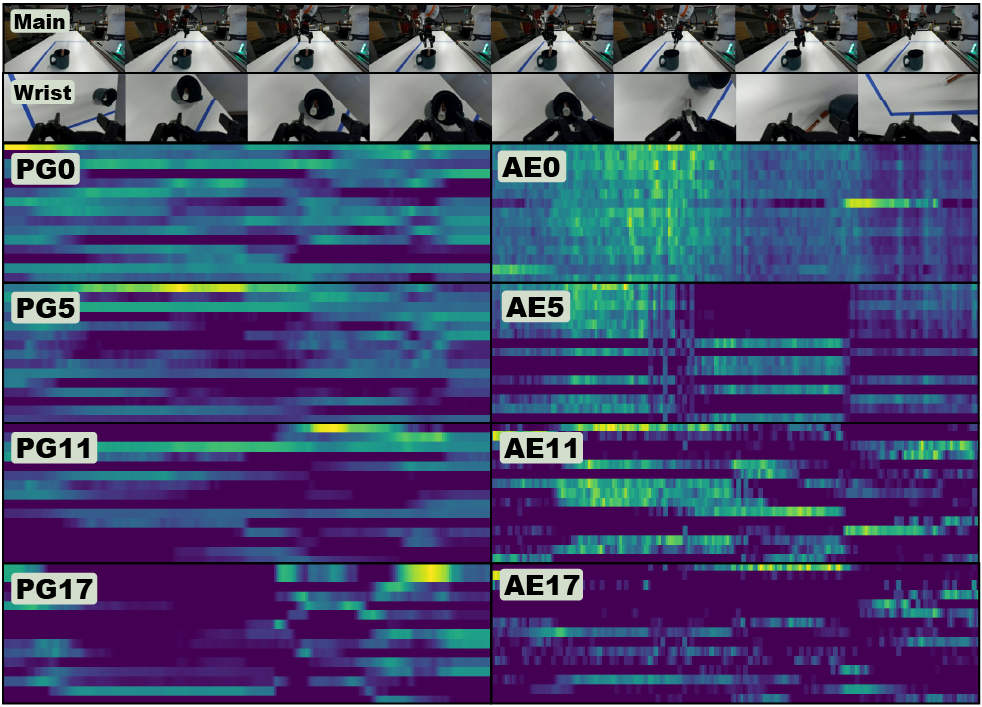}
        \caption{DROID episode 855: \textit{Put the glass lid on the black pot}.}
        \label{fig:top15_droid}
    \end{subfigure}
    \hfill
    \begin{subfigure}[t]{0.32\linewidth}
        \centering
        \includegraphics[width=\linewidth]{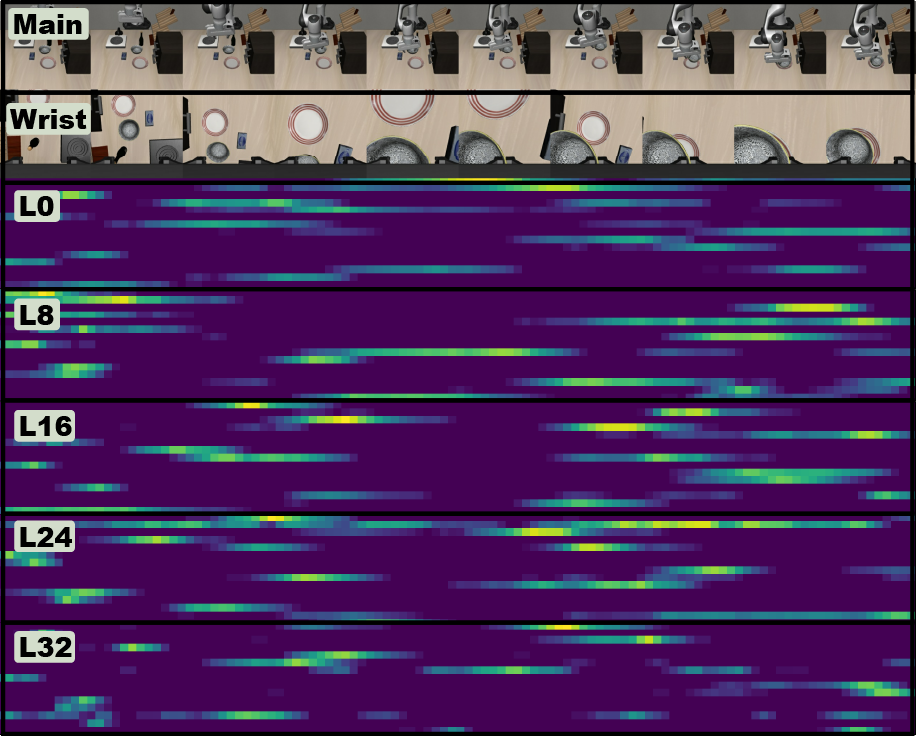}
        \caption{OpenVLA episode 0: \textit{put the bowl on the plate}.}
        \label{fig:top15_openvla}
    \end{subfigure}
    \caption{Top 15 most active SAE features across layers for single episodes from LIBERO (left), DROID (middle), and OpenVLA (right).}
    \label{fig:top15_threeway}
\end{figure}

\Cref{fig:top15_threeway} displays the top 15 SAE features per layer for a single episode from $\pi_{0.5}$ LIBERO, $\pi_{0.5}$ DROID and OpenVLA LIBERO-Object, providing a qualitative overview of how activation structure varies across layers, architectures, and datasets.

For $\pi_{0.5}$, layers PG5 and PG11 contain the most salient and interpretable features. PG0, as the first transformer layer, produces activations that closely resemble the input embeddings; comparing \Cref{fig:top15_libero} PG0 to the SigLIP embedding baseline in \Cref{fig:embed_baseline} confirms this similarity. In contrast with the early PG layers, PG17 contains patterns that more closely mirror the action expert layers. This is consistent with $\pi_{0.5}$'s knowledge-insulation training, in which the VLM backbone produces action tokens in addition to the action expert's continuous actions, suggesting that the later PaliGemma layers have a similar action-encoding structure to the action expert. The action expert layers display a qualitatively distinct pattern. SAE features are activated in distinct phases of manipulation, punctuated by grasp events. We interpret these as motion primitives encoding characteristic directions of motion. This motion-primitive-like structure is consistent across both the LIBERO and DROID action expert layers (Fig. \ref{fig:top15_libero} and Fig. \ref{fig:top15_droid}), despite the two models relying on entirely different action representations (end-effector pose for LIBERO versus joint positions for DROID). 

Comparing the two $\pi_{0.5}$ variants, DROID activations are noticeably denser than their LIBERO counterparts. We attribute this to the substantially greater visual diversity of DROID's real-world scenes relative to LIBERO's simulated environments, which requires the SAE to recruit more features simultaneously to represent each timestep. The OpenVLA features in Fig.~\ref{fig:top15_openvla} show a qualitatively different texture from $\pi_{0.5}$. As OpenVLA is just a VLM containing no action expert, its activation patterns more closely resemble the PaliGemma backbone layers than the action expert layers. Although individual features appear sparser in the heatmap, the fraction of features classified as general is in fact substantially lower than in $\pi_{0.5}$ ( \Cref{tab:feature_classification}). However, this lack of generality is to be expected given the relatively small training dataset as discussed in \Cref{sec:method:dataset}.

\subsection{Embedding Baseline}
\begin{figure}[h]
    \centering
    \includegraphics[width=0.5\linewidth]{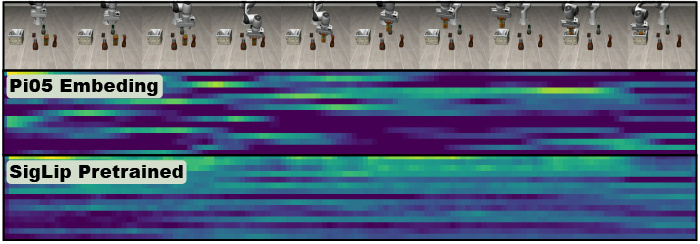}
    \caption{Top 15 SAE features for the $\pi_{0.5}$ embedding layer (top) and the pretrained SigLIP encoder (bottom) for a single episode, \textit{Pick up the orange juice and place it in the basket}. The SigLIP encoder has not been exposed to any robotics data.}
    \label{fig:embed_baseline}
\end{figure}

To ensure that the features we identify arise from robotic fine-tuning rather than from existing features in the pretrained vision encoders, we train SAEs on two baselines: the $\pi_{0.5}$ embedding layer and the frozen pretrained SigLIP encoder that has never seen any robotics data. \Cref{fig:embed_baseline} shows the top 15 features for a single episode from each. The $\pi_{0.5}$ embedding activations bear some resemblance to those found in PG0, which is unsurprising given that the two are separated by only a single attention layer and MLP. In contrast, the pretrained SigLIP model produces a qualitatively different and much denser activation pattern, lacking the event-locked structure characteristic of the general features we identify in later layers. This suggests that the interpretable features reported in our main results are not inherited from the pretrained visual representations but instead emerge during robotic fine-tuning.

\section{Steering Experimental Details}
\label{app:steering_details}

\subsection{Experimental DROID Setup}
\label{sec:apendix:experimental_setup}
\begin{figure}[h]
    \centering
    \includegraphics[width=0.90\linewidth]{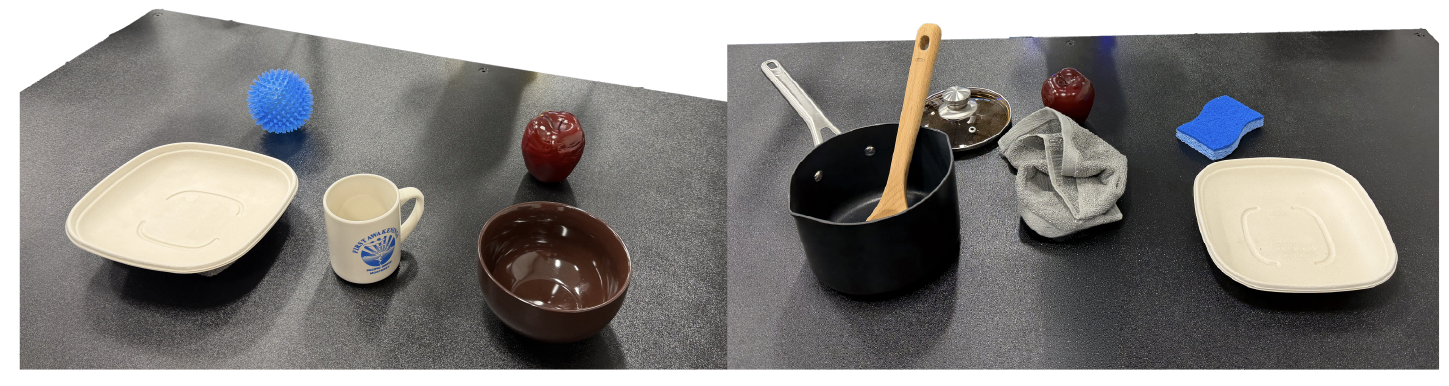}
    \caption{Our experimental setup consists of two scenes shown on the left and right. For each scene several tasks are composed with the available objects, shown in \Cref{tab:real_world_tasks}.}
    \label{fig:real_world_scenes}
\end{figure}

For our DROID experiments we utilize two scenes meant to represent \textit{office} and \textit{kitchen} environments. These scenes are shown in \Cref{fig:real_world_scenes}. The corresponding tasks are shown in \Cref{tab:real_world_tasks}. During inference we run a horizon on 8 actions open loop each time the policy is queried with the current observation. We use the same experimental setup across all experiments, including robot position, camera position, objects and lighting.

\begin{table}[h]
    \centering
    \small
    \begin{tabular}{p{0.44\linewidth} p{0.44\linewidth}}
        \toprule
        \textbf{Scene 1 Tasks} & \textbf{Scene 2 Tasks} \\
        \midrule
        T1. Cover the pot with the lid & T4. Pick up the red apple and put it on the plate \\
        T2. Put the sponge on the plate & T5. Put the blue toy on the plate \\
        T3. Pick up the grey towel and put it on the plate & T6. Pick up the mug and put it on the plate \\
        \bottomrule
    \end{tabular}
    \smallskip
    \caption{Real-world tasks used across the two evaluation scenes.}
    \label{tab:real_world_tasks}
\end{table}

\subsection{Steering Implementation Details}
\label{sec:method:steering}
We expand on the two steering interventions introduced in \Cref{sec:feature_steering}. Given a trained SAE with decoder weight matrix $\mathbf{W}_\text{dec}$, the steering vector for feature $i$ is the $i$-th decoder column,
\begin{equation}
    \mathbf{v}_i = \mathbf{W}_\text{dec}[:, i] \in \mathbb{R}^d,
\end{equation}
where $d$ is the residual-stream dimension at the hooked layer: $d=2048$ for the PaliGemma backbone, $1024$ for the $\pi_{0.5}$ action expert, and $4096$ for OpenVLA. Decoder columns are constrained to unit norm during training, so the magnitude of the additive intervention is controlled by the scalar $\alpha$ alone.

For both the ablative and additive interventions the same perturbation is broadcast over every token position in the sequence and applied at every iteration of the $\pi_{0.5}$ action-expert flow-matching denoising loop. This differs from standard LLM steering~\cite{templeton2024scaling}, where the perturbation is typically once per autoregressive output token.

\subsection{Baseline Implementation Details}
\label{sec:tomlin_implementation}

For the FFN-neuron steering baseline of \citet{mech_interp_tomlin}, we derive steering directions from the MLP down-projection matrix of PaliGemma layer~5. For each target object (sponge, towel, toy, mug) we select the six columns of the down-projection matrix whose top LM-head-projected tokens have the largest overlap with the corresponding object-keyword set (e.g., $\{$``sponge''$\}$ for the sponge task). These six vectors are added to the residual stream simultaneously at every token position and at every iteration of the action-expert denoising loop, mirroring the broadcasting convention used for SAE steering. Six is not chosen arbitrarily and instead follows the experimental protocol in \cite{mech_interp_tomlin}.

For each task we sweep the per-vector coefficient $\alpha \in \{6, 12.5, 25, 50, 75, 100, 125\}$ and report results for the largest value for which the policy remains stable. This is because for both SAE and FFN features have the strongest steering effect at high alpha. Because \cite{mech_interp_tomlin} steers six vectors simultaneously, its total steering magnitude is substantially larger than that of our method. 

For our gripper steering experiments in \Cref{sec:gripper_steering} we use the same overlap criteria to select features corresponding to the object-keyword set (``open" and ``close"). To aid in fair comparison for this alpha sensitive task we steer at the same alpha between our features and the FFN features, averaging across several individual FFN features to replicate the clusters typically used \cite{mech_interp_tomlin}.

\subsection{Feature Ablation Simulation Results}
\label{sec:libero_ablation}
 \begin{figure}[h]
  \centering
  \includegraphics[width=0.65\linewidth]{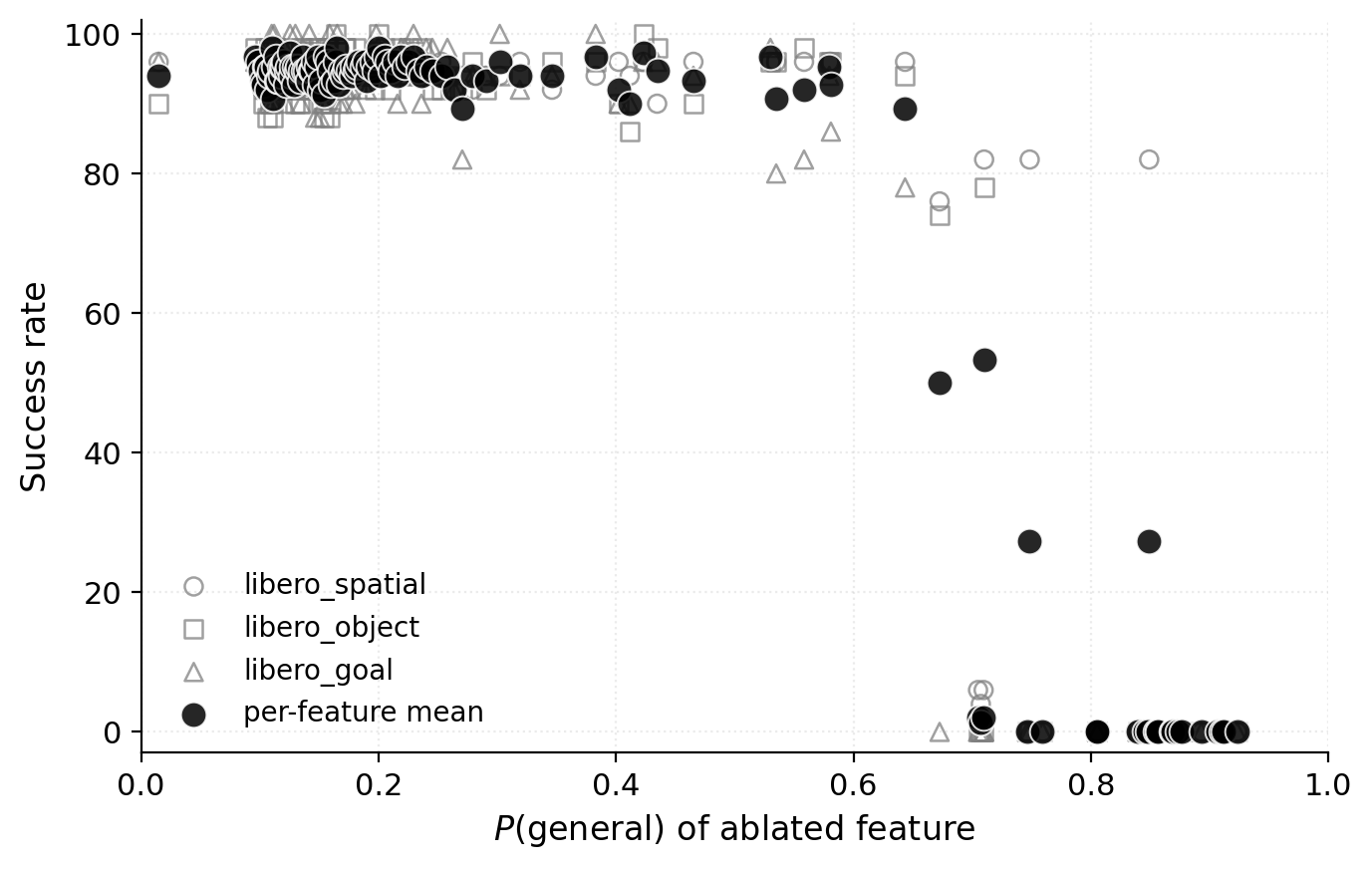}
  \caption{Single-feature ablation on LIBERO. Each black dot is the ablation of a single feature across 150 rollouts. We see a clear pattern that as generality of the ablated feature increases policy performance reduces significantly.}
  \label{fig:libero_project_out}
\end{figure}

We replicate the ablation protocol of \Cref{sec:project_out} in simulation on LIBERO. We use SAEs and classifiers trained on $\pi_{0.5}$ \texttt{(LIBERO)}, fully ablating one feature out of the residual stream for the full episode at each evaluation. We evaluate one task from each of \texttt{libero\_spatial}, \texttt{libero\_object}, and \texttt{libero\_goal} with 50 randomly sampled initial states per (feature, task). We sweep 135 features: 100 selected at evenly spaced ranks plus all 35 features with $P(\text{general}) \geq 0.4$ that were not already included, in order to balance the density of the heavily skewed distribution of features. The full sweep is $20,250$ closed-loop episodes.

\Cref{fig:libero_project_out} shows a similar relationship seen on real-world DROID: features below $P(\text{general}) \approx 0.5$ are indistinguishable from the unsteered policy ($\sim 95\%$), and above the threshold success rate drops sharply, with the four most general features producing 0/600 across all three suites. The classifier's generality score predicts behavioral importance in simulation as cleanly as on hardware.

\subsection{Qualitative DROID Steering Results}
\label{sec:qualitative_steering_results}
In this section we apply all four general features discussed in \Cref{sec:general_features} to steer closed-loop behavior across the tasks and scenes listed in \Cref{tab:real_world_tasks}. Unlike in \Cref{sec:project_out} we use additive steering rather than ablative steering. We steer at $\alpha = 100$ for all timesteps of the episode. We include our general findings below, but encourage readers to look at our \href{https://drvla.github.io}{website} for videos comparing the default behavior to the steered behavior. 

\textbf{F158 (sub-task checkpoint):} At $\alpha{=}100$, the gripper typically dwells over the target object before grasping. This is consistent with our interpretation of F158 as amplifying the sub-task transition signal, which causes the policy to linger in the approach sub-phase rather than commit to grasp acquisition. 

\textbf{F586 (pinch grasp):} At $\alpha{=}100$, the gripper often closes earlier above the object, consistent with this feature encoding a pinch grasp. 

\textbf{F165 (open gripper over target):} At $\alpha{=}100$, the policy executes very similarly to the unsteered policy, with some additional dwelling over the target object before grasping. 

\textbf{F399 (grasp acquisition/placement):} At $\alpha{=}100$, the policy exhibits a pronounced approach-and-retry cycle. It often comes close to completing a grasp, then abruptly lifts upward as if the grasp has already succeeded when it has not. This behavior suggests that the feature can cause the model to behave as though grasp acquisition is complete, prompting an early transition into the transport phase.

Overall, we find that the steering results for these features reinforce our findings in \Cref{sec:general_features}, based on the feature search and activation viewer investigations. One observation that holds throughout our experiments is the impressive robustness of VLA models to activation-level perturbations. Even when steering with a single feature at high magnitude, the policy continues to exhibit goal-directed behavior, generally attempting to approach relevant objects, grasp, or pursue alternative subtasks rather than simply producing degenerate or random motion.

\section{Per-Token SAEs}
\label{app:per_token_saes}

\begin{figure}[h]
    \centering
    \includegraphics[width=0.95\linewidth]{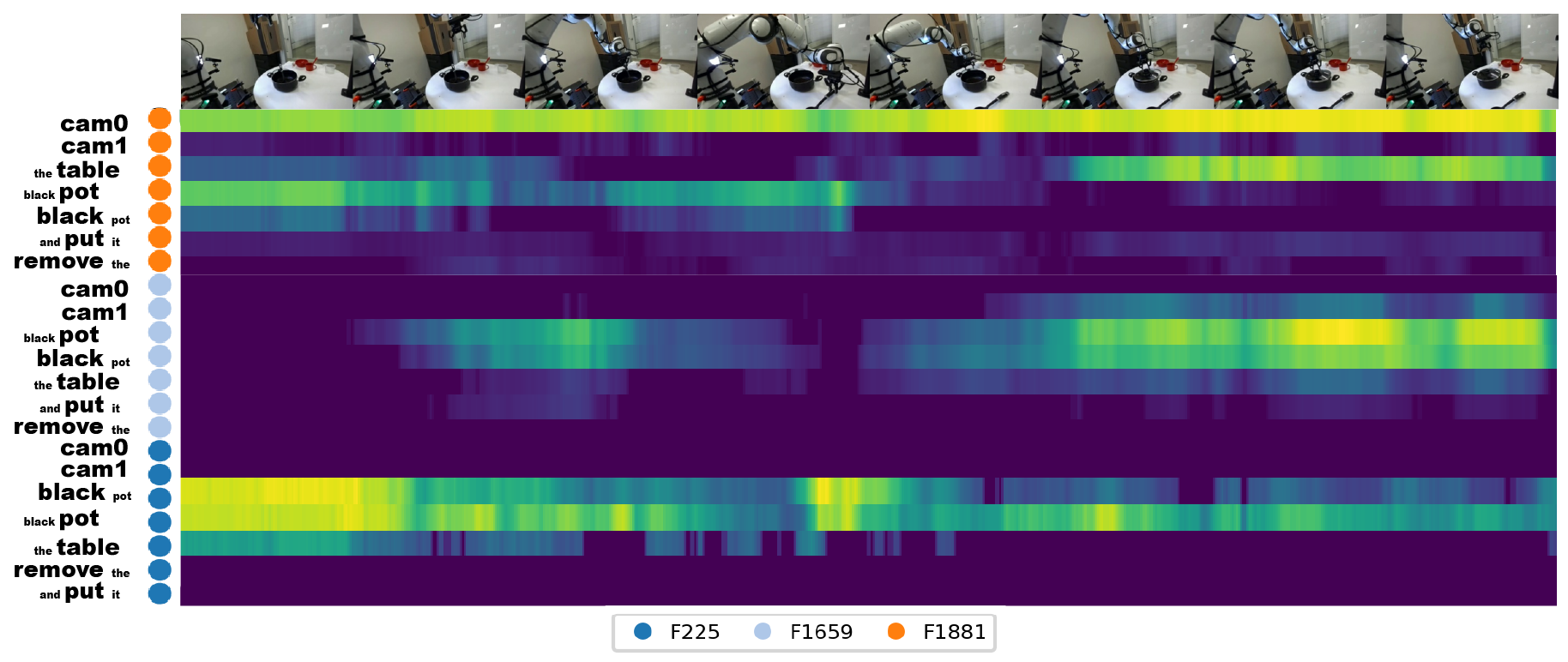}
    \caption{Per-token activations for three features: F225 F1659 and F1881 for $\pi_{0.5}$ DROID PG5. These activations are collected for episode 90 \textit{Remove the black ladle from the black pot and put it on the table, put the lid on the black pot}. We show the SAE activations for 7 tokens: cam0, cam1, table, pot, black, put, and remove. Additional tokens are shown only for context.}
    \label{fig:Ep90_PerToken}
\end{figure}

As discussed in \Cref{app:activation_collection}, we also collect per-token activations. While our summed-token activations reveal several exciting, broadly general features, probing visual–textual semantic alignment requires token-level resolution. To make the problem more tractable, we aggregate all image inputs into a single ``image” token, while we collect individual activations for each ``text" token. We train a single SAE to reconstruct all of these activations. We find that while per-token SAEs provide more insight into the model, more investigation and likely more training data are needed to fully evaluate them. Overall, per-token SAEs are promising but currently less interpretable than their summed-token counterparts. \Cref{fig:Ep90_PerToken} shows our results. We discuss several promising features below.

F1881 is prominent across all episodes but lacks human interpretability. In this episode, it is strongly and nearly uniformly activated on the main camera token and more sparsely, at a lower magnitude, on the wrist camera. All highlighted text tokens exhibit non-zero activation, consistent with a general visual-semantic alignment role. The noun ``pot" and its modifier ``black" display nearly identical activation patterns, suggesting this feature links co-referent tokens within a scene. However, the temporal structure of these activations is not clearly interpretable.

F225 is similarly prominent across episodes but appears to capture purely textual semantic structure. It has zero activation on both camera tokens but strong activation on the nouns in the task instruction. It does not activate on verbs, a pattern consistent across episodes. The tokens ``black" and ``pot" again co-activate, peaking when the wrist camera is positioned over the pot. 

F1659 is an interpretable, memorized feature: its top camera-token activations come exclusively from episodes containing lids and pots, peaking when the lid is placed on the pot. Its top text-token activations correspond to ``pot" or the adjacent space when ``pot" is the final text-token. Within this episode, ``pot" is the highest activating token, again co-activated with ``black" and the wrist camera. The remaining tokens show qualitatively similar but substantially weaker activation. Analogous features in other episodes exhibit strong visual-textual alignment while remaining task-specific. 

Compared with the summed-token SAEs, these per-token SAEs are substantially less interpretable in the general case and require further investigation to fully characterize their visual-semantic alignment properties. Evidence of generality exists but is concentrated in fewer than ten features per layer, and the top-activating text-tokens of these general features are often function words (``the," ``and") rather than semantically informative content words.

\end{document}